\documentclass[letterpaper, 10 pt, conference]{ieeeconf}
\IEEEoverridecommandlockouts                              
\usepackage{times}
\usepackage{epsfig}
\usepackage{graphicx}
\usepackage{multirow}
\usepackage{multicol}

\usepackage{comment}
\usepackage{url}

\usepackage{graphicx}
\usepackage{epsfig}
\usepackage{epic,eepic}
\usepackage{times}
\usepackage{mathptmx}
\usepackage{cite}
\usepackage{color}

\usepackage{times}

\definecolor{red}{rgb}{1,0,0}
\definecolor{green}{rgb}{0,1,0}
\definecolor{blue}{rgb}{0,0,1}
\definecolor{violet}{rgb}{1,0,1}
\definecolor{cyan}{cmyk}{1,0,0,0}
\definecolor{magenta}{cmyk}{0,1,0,0}
\definecolor{yellow}{cmyk}{0,0,1,0}

\definecolor{white}{rgb}{1,1,1}

\usepackage{arydshln}

\newcommand{\CO}[1]{}

\newcommand{\CommentOut}[1]{}

 \newcommand{\editage}[1]{}

\begin{document}

\newcommand{\FIG}[3]{
\begin{minipage}[b]{#1cm}
\begin{center}
\includegraphics[width=#1cm]{#2}\\
{\scriptsize #3}
\end{center}
\end{minipage}
}

\newcommand{\FIGU}[3]{
\begin{minipage}[b]{#1cm}
\begin{center}
\includegraphics[width=#1cm,angle=180]{#2}\\
{\scriptsize #3}
\end{center}
\end{minipage}
}

\newcommand{\FIGm}[3]{
\begin{minipage}[b]{#1cm}
\begin{center}
\includegraphics[width=#1cm]{#2}\\
{\scriptsize #3}
\end{center}
\end{minipage}
}

\newcommand{\FIGR}[3]{
\begin{minipage}[b]{#1cm}
\begin{center}
\includegraphics[angle=-90,width=#1cm]{#2}
\\
{\scriptsize #3}
\vspace*{1mm}
\end{center}
\end{minipage}
}

\newcommand{\FIGRpng}[5]{
\begin{minipage}[b]{#1cm}
\begin{center}
\includegraphics[bb=0 0 #4 #5, angle=-90,clip,width=#1cm]{#2}\vspace*{1mm}
\\
{\scriptsize #3}
\vspace*{1mm}
\end{center}
\end{minipage}
}

\newcommand{\FIGCpng}[5]{
\begin{minipage}[b]{#1cm}
\begin{center}
\includegraphics[bb=0 0 #4 #5, angle=90,clip,width=#1cm]{#2}\vspace*{1mm}
\\
{\scriptsize #3}
\vspace*{1mm}
\end{center}
\end{minipage}
}

\newcommand{\FIGpng}[5]{
\begin{minipage}[b]{#1cm}
\begin{center}
\includegraphics[bb=0 0 #4 #5, clip, width=#1cm]{#2}\vspace*{-1mm}\\
{\scriptsize #3}
\vspace*{1mm}
\end{center}
\end{minipage}
}

\newcommand{\FIGtpng}[5]{
\begin{minipage}[t]{#1cm}
\begin{center}
\includegraphics[bb=0 0 #4 #5, clip,width=#1cm]{#2}\vspace*{1mm}
\\
{\scriptsize #3}
\vspace*{1mm}
\end{center}
\end{minipage}
}

\newcommand{\FIGRt}[3]{
\begin{minipage}[t]{#1cm}
\begin{center}
\includegraphics[angle=-90,clip,width=#1cm]{#2}\vspace*{1mm}
\\
{\scriptsize #3}
\vspace*{1mm}
\end{center}
\end{minipage}
}

\newcommand{\FIGRm}[3]{
\begin{minipage}[b]{#1cm}
\begin{center}
\includegraphics[angle=-90,clip,width=#1cm]{#2}\vspace*{0mm}
\\
{\scriptsize #3}
\vspace*{1mm}
\end{center}
\end{minipage}
}

\newcommand{\FIGC}[5]{
\begin{minipage}[b]{#1cm}
\begin{center}
\includegraphics[width=#2cm,height=#3cm]{#4}~$\Longrightarrow$\vspace*{0mm}
\\
{\scriptsize #5}
\vspace*{8mm}
\end{center}
\end{minipage}
}

\newcommand{\FIGf}[3]{
\begin{minipage}[b]{#1cm}
\begin{center}
\fbox{\includegraphics[width=#1cm]{#2}}\vspace*{0.5mm}\\
{\scriptsize #3}
\end{center}
\end{minipage}
}

\title{\LARGE \bf
Detection and Classification of Pole-like Landmarks for Domain-invariant 3D Point Cloud Map Matching
}

\author{%
Sun Yifei, 
Li Dingrui,
Ye Minying,
Tanaka Kanji\\
University of Fukui, Fukui, Japan \\
\{mf230323, mf210202, yymm2280, tnkknj\}@g.u-fukui.ac.jp
\thanks{The authors are with Department of Engineering, 
University of Fukui, Japan. {\tt\small tnkknj@u-fukui.ac.jp}}}

\maketitle


\newcommand{\eqnA}{
\begin{equation}
S_i^{(t)}
=
\{
x_p:
|| x_p - m_i^{(t)} ||^2
\le
||x_p-m_j^{(t)} ||^2,
\forall j,
1\le j\le k
\}
\}
\end{equation}
}

\newcommand{\eqnB}{
\begin{equation}
m_i^{(t+1)}
=
\frac{1}{s_i^{(t)} 
\lor 
\sum_{x_j\in S_i^{(t)}x_j}
}
\end{equation}
}

\newcommand{\tabA}{
\begin{table}
\caption{Performance results.}\label{tab:A}
\begin{center}
\begin{tabular}{|l|l|l|l|l|}
\hline
\multirow{3}{3em}{Test Dataset} & \multicolumn{4}{l|}{Accuracy} \\
\cline{2-5}
& \multicolumn{2}{l|}{Threshold: 5 m} & \multicolumn{2}{l|}{Threshold: 1 m} \\
\cline{2-5}
& Ours & Baseline & Ours & Baseline \\
\hline
2012-03-17 & 97.66 \% & 92.19 \% & 55.69 \% & 47.51 \% \\
\hline
2012-05-11 & 97.93 \% & 91.37 \% & 48.65 \% & 37.16 \% \\
\hline
2012-11-04 & 96.45 \% & 88.39 \% & 51.08 \% & 43.64 \% \\
\hline
\end{tabular}
\end{center}
\end{table}
}

\newcommand{\figA}{
\begin{figure}
\begin{center}
\hspace*{1cm}\FIG{8}{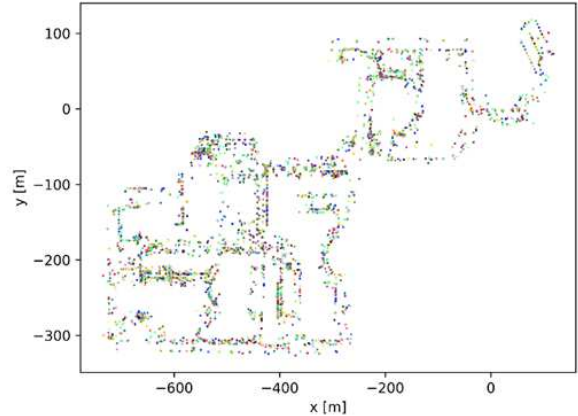}{}
\vspace*{5mm}
\end{center}
\caption{Pole landmark map generated from NCLT dataset. Different color corresponds to different pole class.}\label{fig:A}
\end{figure}
}

\newcommand{\figB}{
\begin{figure}
\begin{center}
\hspace*{1cm}\FIG{8}{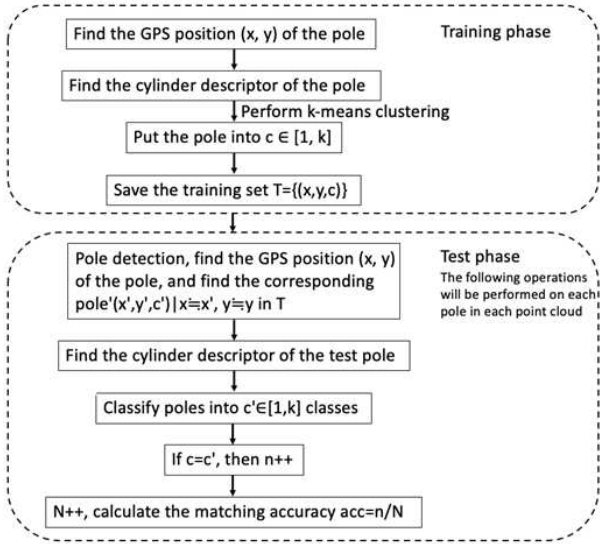}{}
\vspace*{5mm}
\end{center}
\caption{Flowchart for generating hypothesis and scoring by RANSAC.}\label{fig:B}
\end{figure}
}

\newcommand{\figC}{
\begin{figure}
\begin{center}
\hspace*{1cm}\FIG{8}{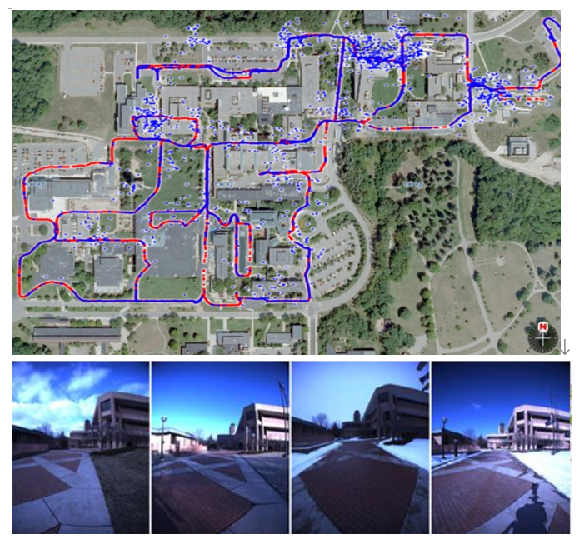}{}
\vspace*{5mm}
\end{center}
\caption{Experimental environment with example results of pole classification (dataset: 2012-03-17).}\label{fig:C}
\end{figure}
}

\begin{abstract}
In 3D point cloud-based visual self-localization, pole landmarks have a great potential as landmarks for accurate and reliable localization due to their long-term stability under seasonal and weather changes. In this study, we aim to explore the use of recently developed deep learning models for pole classification in the context of pole landmark-based self-localization. Specifically, the proposed scheme consists of two main modules: pole map matching and pole class matching. In the former module, local pole map is constructed and its configuration is compared against a precomputed global pole map. An efficient RANSAC map matching is employed to achieve a good tradeoff between computational efficiency and accuracy. In the latter pole class matching module, the local and global poles paired by the RANSAC map-matching are further compared by means of pole attribute class. To this end, a predefined set of pseudo pole classes is learned via k-means clustering in a self-supervised manner. Experiments using publicly available NCLT dataset showed that the pole-like landmark classification method has an improved effect on the visual self-localization system compared with the baseline method.
\end{abstract}


\section{Introduction}

Accurate and reliable localization and map building in the face of various outdoor environments is a fundamental problem for mobile robots and autonomous vehicles. At present, the positioning system based on global satellite navigation system is easily affected by observation conditions such as terrain and weather because the positioning accuracy of the satellite system are often unreliable due to confusing signals and satellite occlusions. Positioning based on dense maps such as grid maps, point clouds or polygon grids is more demanding on memory for negative, dense maps and cannot cope with larger scale map-based self-localization. Furthermore, landmark maps can reduce memory occupation by several orders of magnitude while ensuring positioning accuracy.

Vision is the primary sensor for many localization and place recognition algorithms. We chose a method based on visual self-localization of polar landmarks in a 3D point cloud that relies on polar landmarks extracted from mobile LiDAR data. Pole landmarks appear as part of street lights, traffic signs and tree trunks (Fig. \ref{fig:A}). In addition, pole landmarks are widespread and abundantly distributed in urban areas. It is stable, stable over time under seasonal and weather changes, and its geometry is well defined. Therefore, if a static object can be found as a positioning reference in a dynamic environment, the positioning accuracy will be greatly improved.

\figA

3D LiDAR data is superior to 2D image data in dealing with problems such as illumination and background confusion, and the application of 3D LiDAR data is starting to gain popularity. The localizer proposed by Kim et al. \cite{1} learns point cloud descriptors called Scan Context Images (SCI) and performs robot localization on a grid map by formulating the place recognition problem as place classification using convolutional neural networks. In Cabo et al. \cite{2}, Cuenca et al. \cite{3}, Yu et al. \cite{4}, Zheng et al. \cite{5} and Wu et al. \cite{6}, excellent research results were achieved using high-density and high-precision LiDAR data acquired by MLS (Mobile Laser Scanner) to extract urban street facilities. In Schaefer et al. \cite{7} the authors used a visual navigation system with LiDAR to generate environment maps and then extracted pole-like landmarks from the maps. Finally, the effectiveness of global localization is demonstrated by matching the current scan with the extracted landmarks.
Our observations are that pole landmarks trade discriminative power for detection robustness. A pole is represented as a 2D point on the map. Such low-dimensional representations often make it difficult to distinguish similarly positioned poles. We attach the minimum necessary attribute information to the pole. By doing so, we aim to dramatically improve the identifiability while maintaining the robustness and lightness of the pole landmark. Specifically, this attribution task is formulated as the problem of classifying a pole's attributes into one of a predefined set of attribute candidates (Figure \ref{fig:A}).
The proposed scheme consists of two main modules: pole map matching and pole class matching. In the former module, local pole map is constructed and its configuration is compared against a precomputed global pole map. An efficient RANSAC map matching is employed to achieve a good tradeoff between computational efficiency and accuracy. In the latter pole class matching module, the local and global poles paired by the RANSAC map-matching are further compared by means of pole attribute class. To this end, a predefined set of pseudo pole classes is learned via k-means clustering in a self-supervised manner.  In addition, we establish a baseline method for generating map hypotheses through pole-based self-localization only, and match and score the two sets of hypotheses generated to obtain matching accuracy. Experiments using publicly available NCLT dataset showed that the pole-like landmark classification method has an improved effect on the visual self-localization system compared with the baseline method.

\section{Pole Class Matching}

\subsection{Pole Landmark Identification}

The stability of pole-like landmarks under different seasonal and weather conditions has become a quality landmark for series of self-localization studies. In recent years, many authors have addressed specific problems of vehicle localization using pole landmarks extracted from 3D LiDAR scans. Pole extraction is an important component of localization systems, where researchers are interested not only in fitting geometric primitives to the data and determining pole coordinates, but also in accurate segmentation of pole-like landmarks. All methods provide pole extraction systems, and some researchers have proposed composite pole localization systems. Most of the solutions offered in the study include at least the following two components: a pole detector and a landmark-based pose estimator. A complete localization system is formed by the pole extraction module and the landmark localization module. The detector developed by Li \cite{8} et al. was used to segment the space around the LIDAR sensor and calculate the laser reflection count for each voxel. The poles are then assumed to be located at points that exceed the reflection count threshold.

\subsection{
Pole Landmark Extraction
}

The principle of laser distance measurement is based on focused laser imaging and measuring the return time of the laser signal to calculate distances using sensors and digital acquisition devices. When dealing with distant objects, the laser can be adapted to provide imaging and optical signal information by providing a specific wavelength. As the technology continued to advance, it eventually led to the creation of LIDAR (Light Detection and Ranging). As the technology continues to evolve, LIDAR is now quite mature, and numerous LIDAR companies have tested and commercialized it in self-driving cars, advanced driver assistance systems, mapping, robotics, infrastructure and smart city applications.
Schaeffer et al. proposed a method to subdivide space and extract poles based on the number of laser reflections in each voxel. The occupied and available space is explicitly modeled by the start and end points of the scan. The pole extraction module proposed by Schaefer takes as input a set of registered 3D LiDAR scans, allows estimating the width properties of the poles, and outputs the 2D coordinates of the detected pole centers with respect to the ground plane. The pole extraction module requires the construction of a 3D occupancy map of the scan space, and the pole feature detector can regress the generated pole map into a set of detected pole position and width estimates for each voxel.

\subsection{
Pole Landmark Clustering
}

The standard k-means clustering algorithm is described as follows. The computation begins with randomly selecting k=200 observations from the data set and used as initial means. Scattering the initial means and setting up random partitions to place all initial means close to the center of the dataset, the algorithm is performed by alternating between two steps \cite{9}.
Assignment step: assign each observation to the cluster with the closest mean: the cluster with the least square Euclidean distance.
\eqnA
where each is assigned to exactly one, even though it can be assigned to two or more of them.
Update step: recalculate the mean (center of mass) of the observations assigned to each cluster.
\eqnB	

The computation is considered converged and terminated when the allocation no longer changes \cite{10}.

\figB

\section{Pole Map Matching}

Interest in the use of mobile geographic information systems (GIS) in a variety of automotive applications has increased dramatically over the past few years. Car navigation systems use digital maps to guide drivers to their desired destinations. The next generation of driver assistance systems will use enhanced maps to provide precise navigation cues (including locations of speed limits, gas stations, restaurants, etc.) and to assist with vehicle control. In addition, methods to autonomously augment existing maps using video and LiDAR sensors are being developed. The measured vehicle locations are robustly assigned to road segments in the digital map. This process is called map matching. Motion structure has long been a very active area of research in computer vision. The estimation of known structures and motions is highly sensitive to noise and outliers. Therefore, during map matching, it is important to minimize the cost function as close as possible to the true noise and outlier conditions. The random sample consensus (RANSAC) algorithm is a widely used robust estimator that has become the dominant stochastic model validation strategy in computer vision. RANSAC and related hypothesis and validation methods have been applied to map matching, visual place recognition, and many other related studies.

\subsection{RANSAC Algorithm}

The main task of RANSAC is to generate hypotheses and score them. Among them, for the running process of RANSAC in the baseline self-localization system, n poles are selected as input, the distance between any two input poles is calculated to bring into a pre-generated distance lookup table to find a hypothesis that meets the conditions, and after rotating the hypothesis by translational transformation, if any point in the hypothesis matches the position of a point in the global map, the score is increased by 1 point, the cycle iterates through all hypotheses, and the score in the hypothesis with the highest rotation-translation matrix will be selected as the final hypothesis. For the pole-based self-localization system, the location information of the points in the hypothesis in the scoring benchmark is combined with the classification information of the pole-like landmarks. If any point in the hypothesis matches a point position in the global map, or any point with the same pole classification category $c$ as a point classification category c in the global map, then the score is increased by 1, iterating through all hypothesis cycles, the rotation-translation matrix with the highest score in the hypothesis is selected as the final hypothesis for the pole class group. The specific flow chart is shown in Fig. \ref{fig:B}.

\figC

\section{Experiments}

\subsection{Dataset}

We use the NCLT dataset to evaluate the proposed approach: the accuracy of the self-localization system is improved by the joint combination of pole classification and self-localization. The NCLT (North Campus Long Term) \cite{11} dataset was acquired from two rounds of Segway robots provided by the University of Michigan campus (Figure \ref{fig:C}). The NCLT dataset can be used for tasks such as navigation and mapping in changing environments using vision or LiDAR, and is also well suited for robotics research testing large-scale, long-term autonomous problems such as long-term localization in urban environments.
These datasets explore the campus through iterations that include seasonal changes, different weather conditions (e.g., falling leaves and rain and snow), different times of day, indoor and outdoor environments, a large number of static objects (trees, street signs, light poles, etc.), and moving things (e.g., pedestrians, bicycles, and cars), as well as long-term structural changes resulting from two large construction projects that are under constant construction. The pole landmarks in the NCLT dataset can serve as a good reference for the visual localization of poles in this study. Although the trajectories vary considerably between sessions, there is significant overlap in the trajectories. For more detailed information on the NCLT dataset, see \cite{11}.

\subsection{Results}

We record the results of different models on different test datasets and mathematically analyze all the results. For pole classification, the setting of variable $k$ value in K-means algorithm, we calculated the optimal $k$ value for the correct rate of visual position recognition system by repeated experiments. 

We use two different error thresholds: 5 meters and 1 meter. When the error between the predicted positioning $(x^{pred}, y^{pred})$ and the real positioning $(x, y)$ is smaller than the threshold, the self-localization is regarded as successful. We computed the percentage of accuracy, which is the number of successful test images normalized by the total number of test images.

After that, we tested other datasets and obtained the pole classification accuracy of the proposed and baseline self-localization frameworks as the test results. The results of the pole classification accuracy with the proposed framework and the accuracy of the baseline framework are shown in Table \ref{tab:A}.

\tabA

A visual aerial view of the matching of the pole classification with the proposed and baseline self-localization frameworks are shown in Fig. \ref{fig:C}. In Fig. \ref{fig:C}, the red line is the original trajectory localization point and the blue line is the predicted localization point. If the localization is successful, the points of the two colors are close to each other. The higher the localization accuracy, the higher the overlap between the red and blue lines. By presenting such a visualization, readers can visually judge the success rate of self-localization.

It is clear from the results that our proposed 
pole-classification -based self-localization
outperforms the baseline model in terms of position recognition accuracy and overall model robustness, which validates the gain of the pole classification -based position recognition system.

\section{Concluding Remarks}

In this paper, we propose a new method to obtain point cloud information using a pole extractor, perform a columnar representation of the point cloud information features and classify the pole-like landmarks using a K-means clustering algorithm, and then use the RANSAC model as a decision module to generate matching hypotheses and set a baseline method as a comparison to achieve accurate and effective visual self-localization.
We use the method of dividing the LiDAR point cloud into cylindrical partitions to generate more balanced points for the cylindrical partitions, and then obtain pole-like landmark features based on the partitions to reduce the dimension of the input data, and the subsequent classification difficulty. We use K-means clustering algorithm to classify the obtained pole landmark features. The algorithm principle and process of RANSAC is described in detail using the robust RANSAC parameter estimation algorithm, and the algorithm is applied to the RANSAC generation assumptions and evaluation in the process of visual self-localization research.
We select the NCLT dataset to verify the feasibility and effectiveness of pole-like landmark-based classification in visual place recognition. The stability of the results from multiple test sets is experimentally analyzed by comparing with the baseline method and tuning the parameters with accuracy as the performance metric. The experimental data show that our proposed model achieves excellent performance in terms of self-localization accuracy and overall model robustness compared to the baseline method.

\bibliographystyle{IEEEtran} 
\bibliography{reference}

\begin{thebibliography}{10}
\providecommand{\url}[1]{#1}
\csname url@rmstyle\endcsname
\providecommand{\newblock}{\relax}
\providecommand{\bibinfo}[2]{#2}
\providecommand\BIBentrySTDinterwordspacing{\spaceskip=0pt\relax}
\providecommand\BIBentryALTinterwordstretchfactor{4}
\providecommand\BIBentryALTinterwordspacing{\spaceskip=\fontdimen2\font plus
\BIBentryALTinterwordstretchfactor\fontdimen3\font minus
  \fontdimen4\font\relax}
\providecommand\BIBforeignlanguage[2]{{%
\expandafter\ifx\csname l@#1\endcsname\relax
\typeout{** WARNING: IEEEtran.bst: No hyphenation pattern has been}%
\typeout{** loaded for the language `#1'. Using the pattern for}%
\typeout{** the default language instead.}%
\else
\language=\csname l@#1\endcsname
\fi
#2}}

\bibitem{1}
G.~Kim, B.~Park, and A.~Kim, ``1-day learning, 1-year localization: Long-term
  lidar localization using scan context image,'' \emph{IEEE Robotics and
  Automation Letters}, vol.~4, no.~2, pp. 1948--1955, 2019.

\bibitem{2}
C.~Cabo, C.~Ordo{\~n}ez, S.~Garc{\'\i}a-Cort{\'e}s, and J.~Mart{\'\i}nez, ``An
  algorithm for automatic detection of pole-like street furniture objects from
  mobile laser scanner point clouds,'' \emph{ISPRS Journal of Photogrammetry
  and Remote Sensing}, vol.~87, pp. 47--56, 2014.

\bibitem{3}
B.~Rodr{\'\i}guez-Cuenca, S.~Garc{\'\i}a-Cort{\'e}s, C.~Ord{\'o}{\~n}ez, and
  M.~C. Alonso, ``Automatic detection and classification of pole-like objects
  in urban point cloud data using an anomaly detection algorithm,''
  \emph{Remote Sensing}, vol.~7, no.~10, pp. 12\,680--12\,703, 2015.

\bibitem{4}
Y.~Yu, J.~Li, H.~Guan, C.~Wang, and J.~Yu, ``Semiautomated extraction of street
  light poles from mobile lidar point-clouds,'' \emph{IEEE Transactions on
  Geoscience and Remote Sensing}, vol.~53, no.~3, pp. 1374--1386, 2014.

\bibitem{5}
H.~Zheng, R.~Wang, and S.~Xu, ``Recognizing street lighting poles from mobile
  lidar data,'' \emph{IEEE Transactions on Geoscience and Remote Sensing},
  vol.~55, no.~1, pp. 407--420, 2016.

\bibitem{6}
F.~Wu, C.~Wen, Y.~Guo, J.~Wang, Y.~Yu, C.~Wang, and J.~Li, ``Rapid localization
  and extraction of street light poles in mobile lidar point clouds: A
  supervoxel-based approach,'' \emph{IEEE Transactions on Intelligent
  Transportation Systems}, vol.~18, no.~2, pp. 292--305, 2016.

\bibitem{7}
A.~Schaefer, D.~B{\"u}scher, J.~Vertens, L.~Luft, and W.~Burgard, ``Long-term
  urban vehicle localization using pole landmarks extracted from 3-d lidar
  scans,'' in \emph{2019 European Conference on Mobile Robots (ECMR)}.\hskip
  1em plus 0.5em minus 0.4em\relax IEEE, 2019, pp. 1--7.

\bibitem{8}
F.~Li, S.~Oude~Elberink, and G.~Vosselman, ``Pole-like road furniture detection
  and decomposition in mobile laser scanning data based on spatial relations,''
  \emph{Remote sensing}, vol.~10, no.~4, p. 531, 2018.

\bibitem{9}
D.~J. MacKay and D.~J. Mac~Kay, \emph{Information theory, inference and
  learning algorithms}.\hskip 1em plus 0.5em minus 0.4em\relax Cambridge
  university press, 2003.

\bibitem{10}
J.~A. Hartigan, M.~A. Wong, \emph{et~al.}, ``A k-means clustering algorithm,''
  \emph{Applied statistics}, vol.~28, no.~1, pp. 100--108, 1979.

\bibitem{11}
N.~Carlevaris-Bianco, A.~K. Ushani, and R.~M. Eustice, ``University of michigan
  north campus long-term vision and lidar dataset,'' \emph{The International
  Journal of Robotics Research}, vol.~35, no.~9, pp. 1023--1035, 2016.

\end{thebibliography}

\end{document}